\documentclass[letterpaper]{article}

\usepackage{aaai}
\usepackage{times}
\usepackage{helvet}
\usepackage{courier}
\usepackage{graphicx}
\usepackage{figsize}
\usepackage{multirow}

\title{Online Search Cost Estimation for SAT Solvers}

\author{Shai Haim\and Toby Walsh \\
\{shai.haim,toby.walsh\}@nicta.com.au\\
School of Computer Science and Engineering,\\ 
The University of New South Wales\\
Sydney, Australia \\ ~\and \\
NICTA \\
Sydney, Australia\\
}

\begin{document}
\maketitle

\begin{abstract}

We present two different methods for estimating the cost of solving SAT problems. The methods focus on the online behaviour of the backtracking solver, 
as well as the structure of the problem. Modern 
SAT solvers present several challenges to estimate 
search cost including coping with
non-chronological backtracking, learning and restarts. 
Our first method adapt an existing algorithm for 
estimating the size of a search tree
to deal with these challenges. We then suggest a second method 
that uses a linear model trained on data gathered
online at the start of search. We compare the effectiveness of these two methods using random and structured problems. We also demonstrate that predictions made in early restarts can be used to improve later predictions. We conclude by showing that the cost of solving a set of problems can be reduced by selecting a solver from a portfolio based on such cost estimations.
\end{abstract}

\section{Introduction}

Estimating the cost of solving a $NP$-hard problem like 
propositional satisfiability (SAT)
is a difficult task. Simple backtracking SAT solvers like DPLL
unfold a proper-binary decision tree. 
The Weighted Backtrack Estimate (WBE) \cite{Kilby2006}, 
which is an adaptation of Knuth's offline sampling method \cite{knuth1975}
can generate good estimates of search cost 
for such solvers. However, more modern SAT solvers present several 
challenges for estimating their run-time. 
For instance, clause learning repeatedly changes the problem the solver faces. 
Estimation of the size of the search tree at any point should take into consideration 
the expected changes that future learning clauses will cause. 
As a second example, restarting generates a new search tree which again needs
to be taken into account by any prediction method. 

Our approach to these problems is to use an on-line method to estimate the cost of the search by observing the solver's \emph{behaviour} in a small part of search. Our first method is an extension of an existing method. It adapts
the Weighted Backtrack Estimator \cite{Kilby2006} to support 
non-chronological backtracking. 
Our second method uses machine learning. We show that using machine learning, it is possible to achieve good estimates at a very early stage of the search, by exploiting data gathered from other instances from the same ensemble. These two methods are tied together, since the tree size estimated by the WBE method is a useful feature for the machine learning method.

\section{Background}
\label{Sec:background}

\label{WBE-RE}
Knuth used probing sample to estimate the size of a backtrack search tree
 \cite{knuth1975}.
If $b_{i}$ is the branching rate observed at depth $i$ of the probe,
then $1+b_{1}+b_{1}\cdot b_{2}+\ldots$ is an unbiased estimation for 
the size of the tree. Despite its simplicity, this method is strikingly effective.  {Unfortunately, random probing cannot be directly used during backtrack
search}. Inspired by Knuth's method, 
Kilby et. al. proposed two online methods to estimate the size of a search tree
during backtracking search \cite{Kilby2006}: The Weighted Backtrack 
Estimator, which is discussed in depth in the next section, and the Recursive Estimator. The Recursive Estimator
simply assumes that any unexplored right subtree is identical in size to the left subtree. Both methods are unbiased
and independent of the problem or solver, but since they are both estimating the size of a complete binary search tree, they do not work directly in modern solvers and perform poorly for most satisfiable instances. 
\citeauthor{Kokotov2000} suggested some {similar techniques} to the RE. The Progress Bar for SAT Solvers \cite{Kokotov2000} estimates the remaining time to solving a SAT instance by observing previously visited nodes. The estimate is calculated using either \emph{historical} or \emph{predictive}  heuristics. Historical estimators use the average observed for previous nodes at the same depth.
The \emph{simple average} estimator just uses a straight forward average,
whilst the \emph{weighted average} favours more recent subtrees. 
Predictive estimators, on the other hand, are based on the size
of the subproblem (e.g. number and size of the clauses)

\label{ml-background}

Machine learning has also been used to estimate search cost. By observing 
the solver as it solves the problem, we might be able to estimate 
how long it will take for the solver to solve it.
Horovitz et al  used a Bayesian approach to classify CSP and SAT problems 
according to their runtime \cite{Horvitz2001}. 
Whilst this work is close to that presented here, there are 
some significant differences. For example, they used SATz-Rand, which does not 
have some of the complex features tackled here such as clause learning.
Xu et. al \cite{Xu2007} used machine learning to tune
empirical hardness models \cite{Leyton-Brown2002}. Learning mostly used
static features of the problem instance. The only exception 
was a group of features generated by probing the search space using 
DPLL and stochastic search. Their method gives a probabilistic estimate
of runtime and not, as here, an estimate for a specific run. 
Their search cost estimates were used within
a portfolio based SAT solver \emph{SATzilla} \cite{Xu2007a}.

Finally, an online machine learning method has been developed 
to speed up a QBF solver \cite{Samulowitz2007}. Having solved
different datasets of problems, a multinomial logistic regression
model was built to classify each instance to its best heuristic. 
This model was used to suggest the best heuristic for
new problem instances. Such a technique could also be used dynamically
to change the heuristic used by a solver.

\section{Weighted Backtrack Estimator}
\label{Sec:WBE}
\label{WBE}

We begin by describing how the existing 
WBE algorithm \cite{Kilby2006} can be adapted
to cope with modern SAT solvers. 
At every point in search, the WBE algorithm 
estimates the search tree size as:
\begin{equation}
\frac{\sum^{}_{d\in{D}}prob(d)(2^{d+1}-1)}{\sum^{}_{d\in{D}}prob(d)}
\end{equation}
Where $prob(d)=2^{-d}$ is 
related to the probability that we visit such a depth using random probing,
and D is the multiset of branches lengths visited.
By storing the numerator and denominator, this estimate
can be calculated in constant time and space at every backtrack.
The resulting estimate is unbiased assuming we have a proper binary
search tree.

Since WBE generate a tree size estimation it is significantly more effective for unsatisfiable instances. Moreover, WBE is not
directly applicable to modern SAT solvers
as they perform (conflict driven) backjumping.
By backjumping over nodes, we no longer have a proper binary tree. 
A second problem is that on backtracking to a decision level, 
modern SAT solvers are not forced to branch on the negated decision.
We can instead branch on a new variable. 
Another challenge for WBE is restarts. At every restart point, 
a new tree is generated. Any method to estimate search cost
must take these factors into account.

In order to construct a proper binary search tree representing
the branching decisions of a SAT solver, 
and to compensate for backjumping, we observe 
the two atomic operations performed during search.

\begin{itemize}
\item \emph{assign(v,b):} when $v$ is a variable and $b$ is a Boolean value. This action assigns the variable $v$ the value $b$. This assignment will be kept in the next level of the search stack. After every assignment a unit propagation process takes place. The values that are assigned in this process are also considered to be assigned in this decision level.
\item \emph{backtrack(d):} backtrack back to decision level $d$. Unassign all variables assigned in any decision level equal to or greater than $d$. Any backtrack is also followed by unit propagation.
\end{itemize}

\begin{figure}
\centering
\subfigure[Sequence of actions] 
{
    \label{Fig:wbe:stack}
    \includegraphics[width=5.0cm, height=3.8cm]{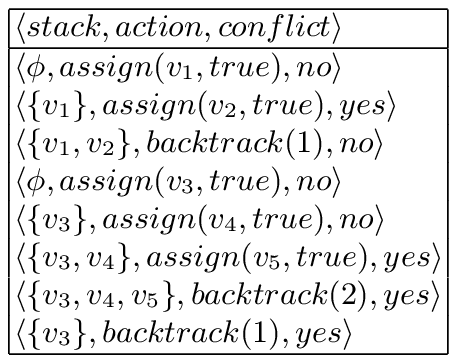}
}
\hspace{1.0cm}
\subfigure[Resulting binary search tree] 
{
    \label{Fig:wbe:after}
    \includegraphics[width=3.0cm, height=3.8cm]{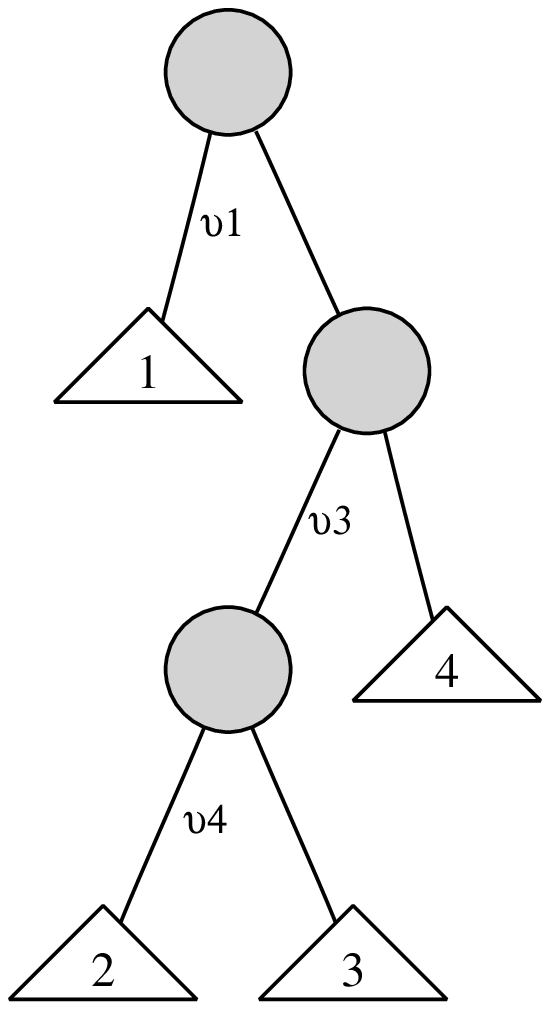}
}\caption{Conversion of a DPLL trace into a binary tree. In \ref{Fig:wbe:stack}, $stack$ is the assignment stack before the action, 
$action$ is the action taken, and $conflict$ denotes if this generates a
conflict. In \ref{Fig:wbe:after}, conflicts are numbered, and
edges labelled with assignments. Since decisions $v_{2}$ and $v_{5}$ are backjumped over, they do not appear as labels.}
\label{Fig:wbe} 
\end{figure}

A binary tree can be generated as follows: we branch left from a node for every $assign$ operation, and we branch right when we $backtrack$ back to the node, 
even if the next assignment is at the same decision level.
If we backjump over node $n$, this node is removed. Note that
node depths in the binary tree no longer correspond with decision levels,
Figure\ref{Fig:wbe} shows an example of this technique. In Figure \ref{Fig:wbe:stack}, we see a list of $\left<stack,action,conflict\right>$ tuples, representing a sequence of actions and the resulted assignment stack. The tree in Figure \ref{Fig:wbe:after} is the explicit proper binary tree corresponding to the same sequence of steps. Note that in both cases there are 4 conflicts, but note that the node depths change.

\section{WBE for Conflict Driven search}

Every time a backjump occurs, 
WBE needs to update the depths of leaves beneath this
backjump. This is not possible if we just store
an accumulated sum for the denominator and numerator
in the WBE estimation. Fortunately, 
the WBE estimation can be computed by observing two different 
parameters which are easy to adjust after backjumping. 
The first, $C$, is a simple counter of the nodes encountered 
so far in an in-order tree search (counting a node only after 
backtracking from its left subtree). The second, $P$, is the partial size 
of the tree explored assuming it is a complete binary tree. 
At any point in search, the WBE estimate
can be generated by calculating:
\begin{equation}
\frac{C}{P} - 1
\end{equation}
Where $C$ is the number of nodes encountered so far and: 
\begin{equation}
P=\frac{1}{\sum^{}_{n \in closed}(2^{d(n)+1})}
\end{equation}
Where $d(n)$ is the depth of node $n$ and $closed$ is the subset of 
nodes in the current branch whose left child has been closed.
We can show this as follows:

\begin{displaymath}\frac{\sum^{}_{d\in{D}}prob(d)(2^{d+1}-1)}{\sum^{}_{d\in{D}}prob(d)}\end{displaymath}\\
\begin{displaymath}=\frac{\sum^{}_{d\in{D}}prob(d)2^{d+1} - \sum^{}_{d\in{D}}prob(d)}{\sum^{}_{d\in{D}}prob(d)}\end{displaymath} \\
\begin{displaymath}=\frac{\sum^{}_{d\in{D}}prob(d)2^{d+1}}{\sum^{}_{d\in{D}}prob(d)} - 1 = \frac{2\left|D\right|}{\sum^{}_{d\in{D}}prob(d)} -1\end{displaymath}\\
\begin{displaymath}=\frac{C}{\sum^{}_{d\in{D}}prob(d)} - 1\end{displaymath} \\

Note that $C = 2\left|D\right|$ as $C$ is increased by 2 for every conflict (once for the leaf and again for the node we backtrack to).
Finally, we can show by induction on the depth of the tree that:
\begin{equation}
\sum^{}_{d\in{D}}prob(d) = \frac{1}{\sum^{}_{n \in closed}(2^{d(n)+1})} = P
\end{equation}
where $d(n)$ is the depth of node $n$ and $closed$ is 
the subset of the nodes in the current branch whose left child has
been closed. 

%
%
\label{WBE:complexity}
Both $C$ and $P$ can be computed incrementally as we 
branch and backjump. Since the search tree is not kept
explicitly in memory, $closed$ is computed using a bit array.
This increases the space and time complexity of
calculating WBE by $O(d)$ where $d$ is the maximum
depth. We can avoid increasing the amortized complexity 
if we estimate search cost at only every $O(d)$ nodes.

Restarts create an extra challenge for WBE. 
Upon restarting, a new tree is generated. The search cost
estimation therefore needs to change. Since WBE generate a \emph{tree size} 
estimate, we can generate a cost estimation by adding the tree size 
estimated by WBE to the number of nodes explored until we reach 
a restart big enough to explore such a tree. 

\section{Linear model prediction (LMP)}

\label{Sec:LMP}

To learn from more than just the size of previously
explored search trees, we developed
an online machine learning method.
We estimate the runtime on a problem $ \cal P$$\in E$, when $E$ 
is an ensemble of problems after training a linear model using 
a subset of problems $\cal T $$\subset E$. For every training 
example $t \in \cal T$ a feature vector 
$x_{t}=\left\{{x_{t,1},x_{t,2},\ldots,x_{t,k}}\right\}$ is created
from on observation window of the search tree. 
We selected features by removing the feature with the smallest 
standardised coefficient until no improvement is observed
based on the standard AIC (Akaike Information Criterion). 
We then search for and eliminate co-linear features in the set. 

Using ridge linear regression, we fit our coefficient vector $w$ 
to create a linear predictor $f_{w}\left(x_{i}\right) =w^{T}x_{i}$. 
We chose ridge regression, since it is a quick and simple technique 
for numerical prediction, and it generally yields good results. 
We predict the log of the number of conflicts.
Since the feature vector is computed online, it is important
that it does not add significant cost to search. The feature vector 
therefore only contains features that can be calculated in constant time. 
We define the \emph{observation window} to be that part of the search 
where data is collected. At the end of the observation window,
the feature vector is computed and the model queried for an estimation.

\begin{table}
	\centering

  \begin{tabular}{ | l | c | c | c | c | c | c |}
  \hline
		\multirow{2}{*}{$Feature$} & \multirow{2}{*}{$init$} &\multicolumn{5}{|c|}{\emph{Observation Window}}\\\cline{3-7}
		 &  & $min$ & $max$ & $avg$ & $SD$ &	$last$\\\hline
		$var$ & $\surd$ &  &  &  &  &  \\
		$cls$ & $\surd$ &  &  &  &  &  \\
		$cls/var$ & $\surd$ & $\surd$ & $\surd$ & $\surd$ & $\surd$ & $\surd$ \\
		$var/cls$ & $\surd$ & $\surd$ & $\surd$ & $\surd$ & $\surd$ & $\surd$ \\
		$FBC$ & $\surd$ &  &  & $\surd$ & $\surd$ & $\surd$ \\
		$FTC$ & $\surd$ &  &  & $\surd$ & $\surd$ & $\surd$ \\
		$ACS$ & $\surd$ &  &  & $\surd$ & $\surd$ & $\surd$ \\
		$SD$ &  &  & $\surd$ & $\surd$ & $\surd$ &  \\
		$BSD$&  &  & $\surd$ & $\surd$ & $\surd$ &  \\
		$BS$ & &  & $\surd$ & $\surd$ & $\surd$ &  \\
		$LCS$ &  & $\surd$ & $\surd$ & $\surd$ & $\surd$ &  \\
		$CCS$ &  & $\surd$ & $\surd$ & $\surd$ & $\surd$ &  \\
		$ABB$ &  & $\surd$ & $\surd$ & $\surd$ & $\surd$ &  \\
		$AAB$ &  & $\surd$ & $\surd$ & $\surd$ & $\surd$ &  \\
		$AAB/ABB$  &  & $\surd$ & $\surd$  & $\surd$ & $\surd$ &  \\
		$ABB/AAB$  &  & $\surd$ & $\surd$  & $\surd$ & $\surd$ &  \\
		$LWBE$ &  & $\surd$ & $\surd$ & $\surd$ & $\surd$ & $\surd$ \\
		\hline
  \end{tabular}
  \caption{The feature vector used by linear regression to construct prediction models}
  	\label{Table:featurevector}
\end{table}

The feature vector measures both problem structure and search behaviour. 
Since data gathered at the start of a restart tends to be noisy and 
less useful, we do not open the observation window immediately. 
To keep the feature vector a reasonable size, we use statistical 
measures of various parameters (that is, the minimum over
the observation window, the maximum, the 
mean, the standard deviation and the last value
recorded). The parameters collected 
are the number of variables ($var$), the number of clauses ($cls$), both the variable to clause ratio and its inverse, the fraction of binary and ternary clauses in the clause database ($FBC$ and $FTC$ respectively), the average clause size ($ACS$), the search depth as it appears in the assignment stack ($SD$) and as it appears in the binary tree generated for the WBE calculation($BSD$), the learnt clauses size ($LCC$) and the conflict clause size ($CCS$), the fraction of assigned vars before backtracking ($ABB$) and after backtracking ($AAB$), the ratio between these two features and its inverse, and the log of the WBE prediction ($LWBE$).
The full list of features used is shown in Table \ref{Table:featurevector}.
All the features used can be calculated in constant time and space 
with the exception of the WBE which takes $O(d)$ time and space. 
We therefore only computed WBE every $d$ conflicts where $d$ is the 
depth recorded at the previous estimate.
 
To deal with restarts, we wait until the observation window
is contained within a single restart. In addition, we
exploited estimates from earlier restarts to help improve
later estimates. To do this, we augmented the feature
vector with all the search cost predictions from previous
restarts. 

\section {Experiments}
\label{Sec:results}

We ran experiments with these two methods using MiniSat \cite{Een2003}.
This is a state-of-the-art modern solver, which uses clause 
learning and clause deletion along with an improved 
version of VSIDS for variable 
ordering and a geometrical restart scheme. 
We used a geometrical factor of 1.5, which is the default for MiniSat.
A geometrical factor of 1.2 yielded results of a similar quality.
We used three different distributions of SAT problems.
\begin{itemize}
\item \emph{rand:} An ensemble of 500 satisfiable
and 500 unsatisfiable randomly generated 3-SAT problems
with 200 to 550 variables and a clause-to-var ratio of 4.1 to 5.0. 
\item \emph{bmc:} An ensemble of software verification problems generated 
by CBMC\footnote{http://www.cs.cmu.edu/~modelcheck/cbmc/} based
on a binary search algorithm coded in C. The different examples 
used different array sizes and different number of loop unwindings. 
In order to generate satisfiable problems, a faulty piece of code 
that causes memory overflow was added to the binary search code. 
These problems create a very homogeneous ensemble of problems. We used 250 satisfiable and 250 unsatisfiable problems.
\item \emph{fv:} An ensemble of hardware formal verification problems
distributed by Miroslav 
Velev\footnote{http://www.miroslav-velev.com/sat\_benchmarks.html}. 
These problems were produced by the same technique but not for the 
same underlaying problem, and create an ensemble which is less 
homogeneous than the previous one. We used 56 satisfiable and 68 
unsatisfiable problems.
\end{itemize} 
Since training examples can be scarce, we restricted
the size of our training set to no more than 500 problems. 
For the formal verification problems, we obviously had far less than that.
In the first part of our experiments, when restarts were
turned off, many of the formal verification problems 
were not solved. Our results in this part will only compare 
the other datasets. When restarts were enabled, all three data sets were used.
In all experiments we used 10-fold cross validation, never using the same
instance for both training and testing purposes.
%
We measured the quality of the predictor by observing the percent of 
predictions which are within a certain factor of the 
correct cost (the \emph{error factor}). 
For example, 80\% for error factor 2, denotes that 
for 80\% of the instances, the predicted search
cost was within a factor of 2 of the actual search cost.

We compare our results with the ones obtained by the Progress Bar 
(PB) \cite{Kokotov2000}.
In order to make the comparison possible, we instrumented MiniSat with the Progress Bar. \citeauthor{Kokotov2000} proposed several different heuristics (Constant, Historical-Basic, Historical-Weighted and Clause Count). We observed
similar results with all of these heuristics. We present
results here for the Historical-Weighted heuristic since it performs slightly better for these data sets. We used the progress bar's default settings. 
Note that if the initial search is too deep, the Progress Bar may not 
provide any estimate.

\subsection{Search Without Restarts}

\begin{figure}
\centering

\includegraphics[width=5.8cm, angle=270]{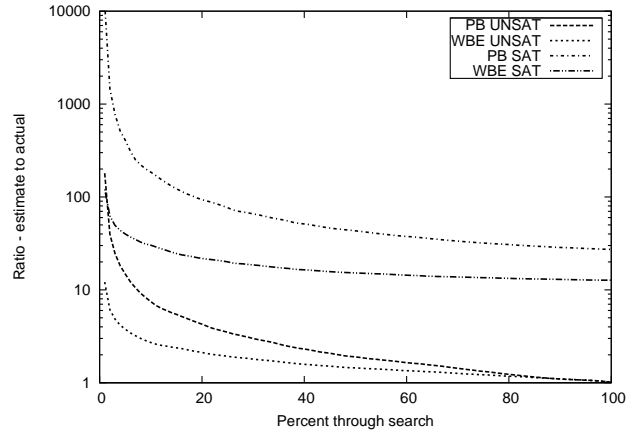}
\caption{Mean ratio of WBE and PB estimates over time for the $rand$ dataset.}
\label{Fig:wbe:time:rand}
\end{figure}

\begin{figure}
\centering

\includegraphics[width=5.8cm, angle=270]{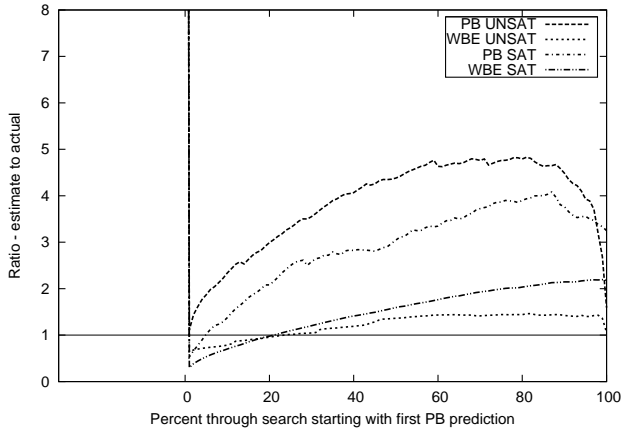}
\caption{Mean ratio of WBE and PB estimates over time for the $bmc$ dataset. Only starts when PB generates its first prediction.}
\label{Fig:wbe:time:bmc}
\end{figure}

Figure \ref{Fig:wbe:time:rand} compares the quality of the WBE
prediction and the Progress Bar prediction over time, for the $rand$ dataset. 
Both predictors return unbiased results for the unsatisfiable problems and 
converge to the correct value given enough time. WBE is generally more accurate than PB both for satisfiable and unsatisfiable instances. In all cases, both estimators start by over-estimating the search cost but their prediction improves with time as we backjump over nodes.
Figure \ref{Fig:wbe:time:bmc} presents the same data for the $bmc$ dataset. For structured problems, WBE initially over estimates search cost
by a large factor (in some cases with by a factor greater than $2^{1000}$).
During this period the Progress Bar does not make any prediction as the
tree is too deep for it to work, and the ``search space left'' is estimated to be 100\%. At some later point in search, we often observed a sharp  improvement in the accuracy of both estimators. Typically this corresponds to search backjumping over an early mistake to a node very close to the root of the tree (or the root itself). For most instances PB starts returning run-time predictions at this point. The WBE also starts returning good prediction at this point.
For unsatisfiable problems in the \emph{bmc} dataset, this point occurs after 72\% of 
the search (on average), but it appears to occur after a smaller percentage 
of the search for harder instances. We found a correlation 
coefficient of $-0.45$ between the total size of the search tree 
and the percent through search where this improvement occurs.

%
%
%

\begin{table}
\centering
\begin{tabular}{ | c | c | c | c | c | c |}
\hline
\multicolumn{3}{| c |}{} & x2 & x4 & x8 \\
\hline
\hline
\multirow{6}{*}{sat} & \multirow{3}{*}{$bmc$} & PB & 	3.8 & 6.8 & 8.9\\
																						&	& WBE & 3.4 & 5.5 & 5.9\\
																						&	& LMP & 40.7 & 68.7 & 85.8\\ \cline{2-6}
										 & \multirow{3}{*}{$rand$} & PB & 	0.9 & 2.6 & 4.9\\ 
																						&	 & WBE & 2.2 & 7.8 & 14.4\\
																						&	 & LMP & 39.7 & 71.3 & 86.6\\\hline\hline
\multirow{6}{*}{unsat} & \multirow{3}{*}{$bmc$} & PB & 	4.9 & 10.3 & 12.8\\
																						&	& WBE & 4.9 & 10.3 & 13.8\\
																						&	& LMP & 36.9 & 68.5 & 93.6\\ \cline{2-6}
										 & \multirow{3}{*}{$rand$} & PB & 	3.7 & 7.4 & 15.5\\
																						&	 & WBE & 12.7 & 29.4 & 47.3\\
																						&	 & LMP & 92.0 & 100.0 & 100.0 \\\hline
\end{tabular}
\caption{Percentage of estimates within error factor after 2000 backtracks}
\label{table:wbe-bp-lmp-2000}
\end{table}

\begin{table}
\centering
\begin{tabular}{ | c | c | c | c | c | c |}
\hline
\multicolumn{3}{| c |}{} & x2 & x4 & x8 \\
\hline
\hline
\multirow{6}{*}{sat} & \multirow{3}{*}{$bmc$} & PB & 	24.5 & 36.2 & 47.3\\
																						&	& WBE & 21.8 & 36.2 & 45.7\\
																						&	& LMP & 49.1 & 78.9 & 95.0 \\ \cline{2-6}
										 & \multirow{3}{*}{$rand$} & PB & 	1.2 & 4.0 & 10.4\\ 
																						&	 & WBE & 4.0 & 12.0 & 22.5\\
																						&	 & LMP & 50.2 & 76.7 & 89.9\\\hline\hline
\multirow{6}{*}{unsat} & \multirow{3}{*}{$bmc$} & PB & 	22.0 & 35.4 & 43.8\\
																						&	& WBE & 32.8 & 48.4 & 48.4\\
																						&	& LMP & 78.1 & 98.4 & 100.0 \\ \cline{2-6}
										 & \multirow{3}{*}{$rand$} & PB & 	17.7 & 42.2 & 58.3\\
																						&	 & WBE & 38.9 & 67.0 & 81.6\\
																						&	 & LMP & 96.7 & 100.0 & 100.0\\\hline
\end{tabular}
\caption{Percentage of estimates within error factor after 35000 backtracks}
\label{table:wbe-bp-lmp-35000}
\end{table}

%

In order to compare the quality of prediction of WBE, PB and LMP, 
we generated an estimate after a constant time, regardless of the true size of the problem. In all cases the estimate generated by LMP was superior to those generated by WBE and PB. Comparisons 
of the performance of those three methods after 2000 and 35000 backtracks are shown in tables \ref{table:wbe-bp-lmp-2000} and \ref{table:wbe-bp-lmp-35000} respectively.
Satisfiable problems are harder to predict for all methods, due to the abrupt way in which search terminates with
open nodes. The linear model deals better with random problems than crafted ones. We conjecture this is due to greater balance in the search tree. WBE performs better than PB for unstructured problems, while they perform similarly for structured instances. The significant improvement of both PB and WBE for structured problems after 35000 backtracks is due to the fact that easier instances are already converging rapidly on the correct answer. \\

\begin{figure}
\centering
    
    \includegraphics[width=5.8cm, angle=270]{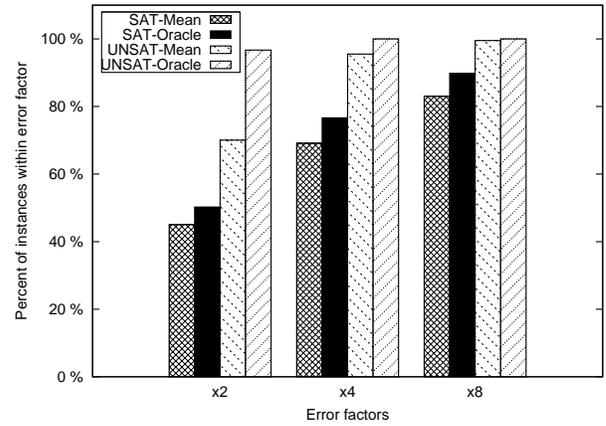}
\caption{Quality of prediction when using an oracle to determine whether an instance is satisfiable or the geometric mean of satisfiable
and unsatisfiable models (denoted \emph{Mean}) - $rand$ dataset}

\label{Fig:weighted:rand}
\end{figure} 

\begin{figure}

\includegraphics[width=5.8cm, angle=270]{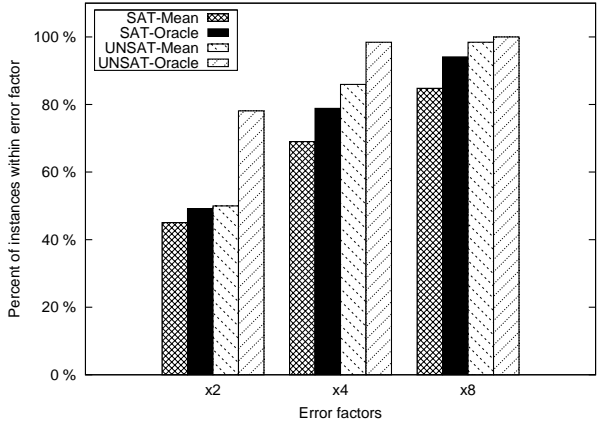}

\caption{Quality of prediction when using an oracle to determine whether an instance is satisfiable or the geometric mean of satisfiable
and unsatisfiable models (denoted \emph{Mean}) - $bmc$ dataset}
\label{Fig:weighted:bmc}
\end{figure}

Since we observe very different behaviour with satisfiable and unsatisfiable 
instances, we trained models on each type of instance separately. 
With a new (non-training) instance, we may not know if it 
satisfiable or unsatisfiable. Indeed, the point of search is often
to decide this. Given a problem of unknown satisfiability,
we therefore queried both models and returned the geometrical mean of the two 
estimates. Figures \ref{Fig:weighted:rand} and \ref{Fig:weighted:bmc} compare using the 
geometric mean of the two models and using an oracle to decide which model to query for the $rand$ and $bmc$ datasets respectively. We see that the geometric mean returns reasonable predictions. 
Alternatively we could train with just one model using both satisfiable and unsatisfiable instances. The performance is similar to the geometric mean of the two models (it is a bit better for $sat$ problems and a bit worse for $unsat$ problems) but is sensitive to the proportion of satisfiable and
unsatisfiable instances.


\subsection{Search With Restarts}

\begin{figure}
\centering
    
\includegraphics[width=5.8cm, angle=270]{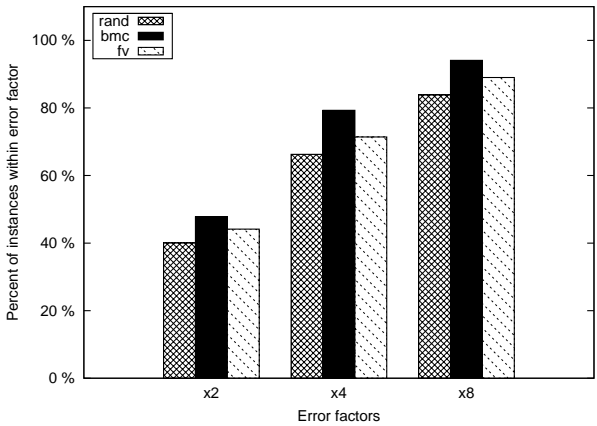}
\caption{Quality of prediction for $sat$ problems with restarts,  (after 2000 backtracks in the \emph{query restart})}
\label{Fig:restarts:sat}
\end{figure}

\begin{figure}
\includegraphics[width=5.8cm, angle=270]{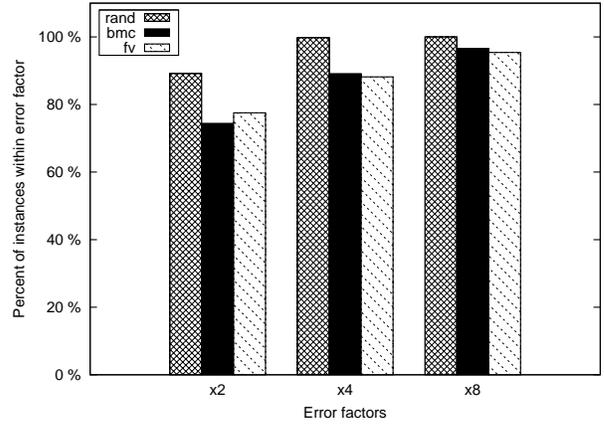}
\caption{Quality of prediction for $unsat$ problems with restarts,  (after 2000 backtracks in the \emph{query restart})}
\label{Fig:restarts:unsat}
\end{figure}

When restarts are used, we have to use smaller observation windows
to give a prediction early in search as many early restarts are small. 
Figures \ref{Fig:restarts:sat} and \ref{Fig:restarts:unsat} compare the quality of prediction of LMP 
for the 3 different datasets, for $sat$ and $unsat$ instances respectively.
The quality of estimates improves with the $bmc$ data set
when restarts are enabled. We conjecture this is a result
of restarts before the observation 
window reducing the noise in the data.

\begin{figure}
\centering
 
\includegraphics[width=5.8cm, angle=270]{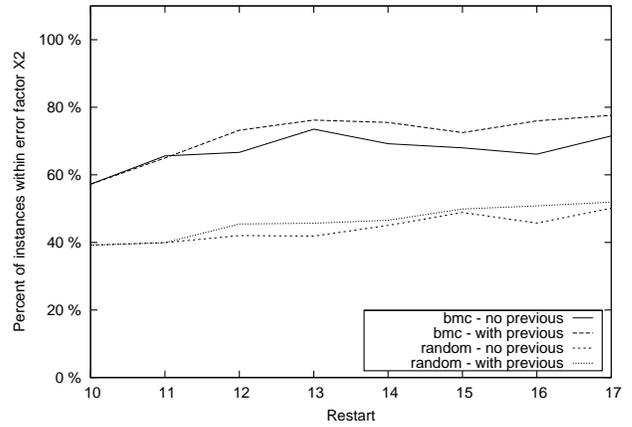}
\caption{The effect of using predictions from previous restarts, for $sat$ instances. Quality of predictions, through restarts, using two datasets (\emph{bmc,rand}). The plots represent the percentage of instances within a factor of 2 from the correct size.}
\label{Fig:rolling:sat}
\end{figure}

\begin{figure}
\includegraphics[width=5.8cm, angle=270]{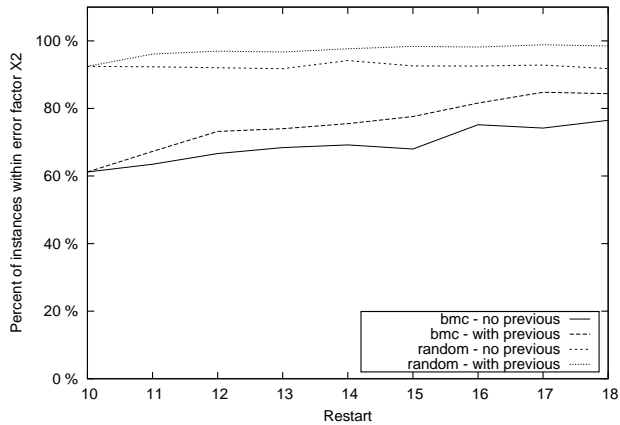}
\caption{The effect of using predictions from previous restarts, for $unsat$ instances. We compare the quality of predictions, through restarts, using two datasets (\emph{bmc,rand}). The plots represent the percentage of instances within a factor of 2 from the correct size.}
\label{Fig:rolling:unsat} 
\end{figure}

In order to check our hypothesis that predictions from previous restarts
improve the quality of prediction in the current restart,
we opened an observation window at every restart. The window 
size is defined by $max(1000,0.01\cdot s)$ and it starts after 
a waiting period of $max(500,0.02\cdot s)$, when $s$ is the 
size of the current restart. At the end of each observation 
window, two feature vectors were created. The first 
$\left(x_{r}\right)$ holds all features from Table \ref{Table:featurevector}, while the second $\left(\hat{x}_{r}\right)$ is defined as
$\hat{x}_{r}=\left\{x_{r}\right\}\cup\left\{f_{w_{1}}\left(x_{1}\right),f_{\hat{w}_{2}}\left(\hat{x}_{2}\right),\dots,f_{\hat{w}_{r-1}}\left(\hat{x}_{r-1}\right)\right\}$. A comparison of the two methods, for $sat$ and $unsat$ instances, is given in Figures \ref{Fig:rolling:sat} and \ref{Fig:rolling:unsat} respectively. We see that predictions from earlier restarts improve
the quality of later predictions but not greatly.

\subsection{Solver selection using LMP}

In our final experiment, we used these estimates
of search cost to improve solver performance. 
We used two different versions of MiniSat. Solver $A$
used the default MiniSat setting (geometrical factor of 1.5), 
while solver $B$ used a geometrical factor of 1.2. 
The challenge is to select which is faster at solving
a problem instance.

\begin{table}

\centering

\begin{tabular}{ | l | l | c | c | c |}
\hline
\multicolumn{2}{| l |}{Dataset} & Oracle & LMP (oracle) & LMP(AVG) \\
\hline
\hline
\multirow{2}{*}{$rand$} & sat 	& 40.8 & 7.0 & 10.5\\
												& unsat & 7.5 & -0.9 & -1.4\\\hline
\multirow{2}{*}{$fv$} 	& sat 	& 66.7 & 17.2 & 16.8 \\
												& unsat & 14.8 & -0.6 & -3.3\\\hline
\multirow{2}{*}{$bmc$} 	& sat 	& 59.6 & 13.3 & 13.6 \\
												& unsat & 17.2 & 0.3 & -0.4\\
\hline
\end{tabular}
\caption{Percentage improvement over average run time for
both solver $A$ and $B$.}
\label{table:portfolio}
\end{table}

Table \ref{table:portfolio} describes the percentage improvement 
of the following strategies compared to the average
run time for both solvers:
\begin{itemize}
\item \emph{oracle:} Use an oracle to tells us which solver is better for the problem ($min(A,B)$).
\item \emph{LMP (oracle):} Use both solvers until each reaches the observation window (restart 9 for solver $A$, restart 19 for solver $B$), and generate a prediction, using an oracle that indicates which model should be queried. Terminate the one predicted to be worse. 
\item \emph{LMP (AVG):} Same as \emph{LMP (oracle)}, but without an oracle to determine whether the instance is \emph{sat} or \emph{unsat}. We instead query both models and use the geometric mean as the prediction.
\end{itemize}
These results show that for satisfiable problems, where solver performance 
varies more significantly, our method reduces the total cost. 
For unsatisfiable problems, where solvers performance does 
not vary as much, our method does not improve search cost.
However, as performance does not change significantly on unsatisfiable instances, 
the overall impact of our method on satisfiable and unsatisfiable
problems is positive. 

\section{Conclusions and Future Work}
\label{Sec:conclusion}
We have presented two different methods to generate 
estimates for the size of the search tree explored by
modern day SAT solvers. The WBE method simply observes
the search tree and requires no prior knowledge of the problem distribution.
This method, like other tree-size based methods performs poorly for satisfiable instances.
The LMP method, on the other hand, uses linear models
which are trained on a problem set. We have shown that it is possible 
to train the model using a relatively small training set, which 
is of value when training examples are in short supply. We have demonstrated the effectiveness of both method on random
problems, as well as on bounded model checking and hardware 
verification problems. We also proposed a simple way to use such 
predictions to select between different SAT solvers. 
There are many directions for future work. For instance, we conjecture it may be
effective to use these methods 
to select between very different types of solver. 
We are currently using LMP to select
between a geometric restart strategy and Luby's restart scheme.

\section*{Acknowledgements}

This paper contains work that is to appear in \cite{Haim2008}. In particular, the LMP method is described in \cite{Haim2008}.
However, the extension of the WBE method to conflict driven solvers, 
along with all results comparing WBE and LMP to the Progress Bar for SAT Solvers are presented here for the first time.\\

The second author is funded by the
Department of Broadband, Communications and the Digital Economy, 
and the Australian Research Council. 

\bibliographystyle{aaai}

\end{document}